\documentclass{article}

\usepackage{microtype}
\usepackage{graphicx}
\usepackage{subfigure}
\usepackage{booktabs} 
\usepackage{times}
\usepackage{epsfig}
\usepackage{graphicx}
\usepackage{amsmath}
\usepackage{amssymb}
\usepackage{textcomp}
\usepackage{multirow}
\usepackage{adjustbox}

\usepackage{multirow}
\usepackage{colortbl}
\usepackage{hhline}
\usepackage{verbatim}

\usepackage{gensymb}
\usepackage{xcolor}
\usepackage{hyperref}

\usepackage{hyperref}

\usepackage[accepted]{icml2021}

\icmltitlerunning{Submission and Formatting Instructions for ICML 2021}

\begin{document}

\twocolumn[
\icmltitle{ST-DETR: Spatio-Temporal\\ Object Traces Attention Detection Transformer}



\icmlsetsymbol{equal}{*}

\begin{icmlauthorlist}
\icmlauthor{Eslam Mohamed Bakr}{to}
\icmlauthor{Ahmad El-Sallab}{to}
\end{icmlauthorlist}

\icmlaffiliation{to}{Valeo R\&D Cairo, Egypt}

\icmlcorrespondingauthor{Eslam Mohamed Bakr}{eslam.mohamed-abdelrahman@valeo.com}
\icmlcorrespondingauthor{Ahmad El-Sallab}{ahmad.el-sallab@valeo.com}

\icmlkeywords{Machine Learning, ICML}

\vskip 0.3in
]



\printAffiliationsAndNotice{\icmlEqualContribution} 

\begin{abstract}
We propose ST-DETR, a Spatio-Temporal Transformer-based architecture for object detection from a sequence of temporal frames. We treat the temporal frames as sequences in both space and time and employ the full attention mechanisms to take advantage of the features correlations over both dimensions.
This treatment enables us to deal with frames sequence as temporal object features traces over every location in the space.
We explore two possible approaches; the early spatial features aggregation over the temporal dimension, and the late temporal aggregation of object query spatial features.
Moreover, we propose a novel Temporal Positional Embedding technique to encode the time sequence information.
To evaluate our approach, we choose the Moving Object Detection (MOD) task, since it is a perfect candidate to showcase the importance of the temporal dimension.
Results show a significant 5\% mAP improvement on the KITTI MOD dataset over the 1-step spatial baseline.
\end{abstract}

\section{INTRODUCTION}
\label{INTRODUCTION}

\begin{figure}
\begin{center}
\centerline{\includegraphics[width=0.9\linewidth]{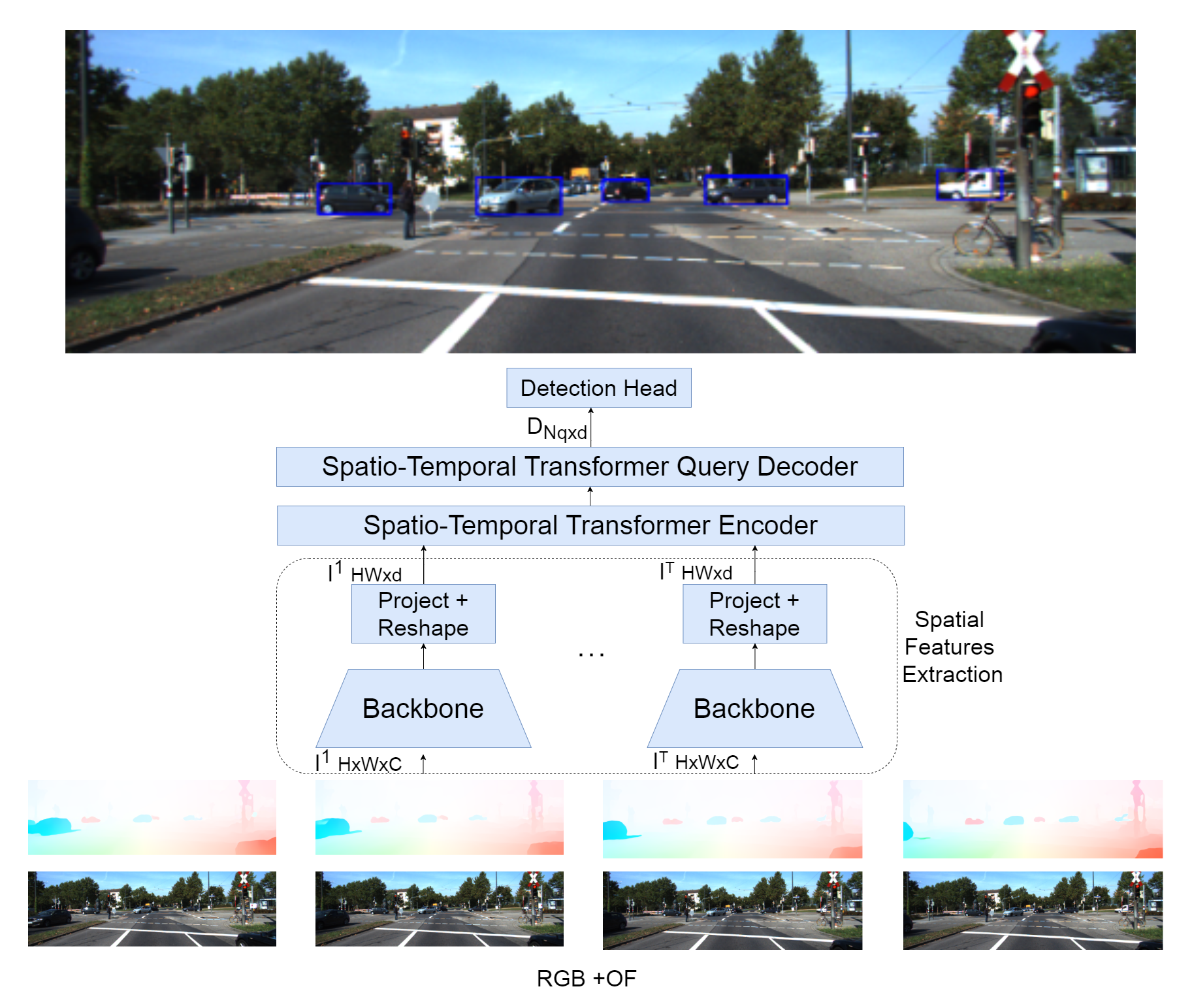}}
\caption{High level framework of the proposed ST-DETR architecture. Lower part shows 4 time-stamp inputs consists of RGB and OF frames for each time-step. Upper part shows the output of the network.}
\label{st-detr}
\end{center}
\end{figure}


For many years, ConvNets \cite{redmon2018yolov3} \cite{mohamed2021insta} have been the architecture of choice in computer vision in general, and for performing object detection tasks in particular. Recently, transformers have shown promising results compared to ConvNets, in object detection, \cite{carion2020end}
,where the input image is treated as a sequence of spatial features, and full attention mechanisms are employed to extract features interactions. This motivates us to extend DETR to handle the sequence information in both spatial and temporal dimensions.

In the general problem setup, we need to perform a Spatio-temporal sequence-to-sequence mapping. The input Spatio-temporal sequence is formulated as a sequence of frames in the temporal dimension within a certain window of time, each is, in turn, a sequence of features in the spatial dimension. The output Spatio-temporal sequence is also a sequence of temporal outputs, each having a list of objects queries.

In order to transform DETR into a Spatio-Temporal model, we undergo some architectural changes. First, we adopt the classical spatial features extraction, and apply it on each input frame across the temporal window. Then we modify both the transformer encoder and decoder to handle the temporal aggregation. Here we have two options, either 1) early temporal aggregation of the spatial features, resulting in a temporal trace of features at each spatial location, or 2) later temporal objects queries aggregation, where we extract the objects queries per time step, and then stack them, resulting in a trace of objects queries. The output of either architecture is a list of object queries features, which are used to predict the bounding boxes and their corresponding classes. The same Hungarian matching and bi-partite loss \cite{carion2020end} are adopted from DETR.


\section{MODEL}

\begin{figure*}[ht!]
\begin{center}
\centerline{\includegraphics[width=0.7\linewidth]{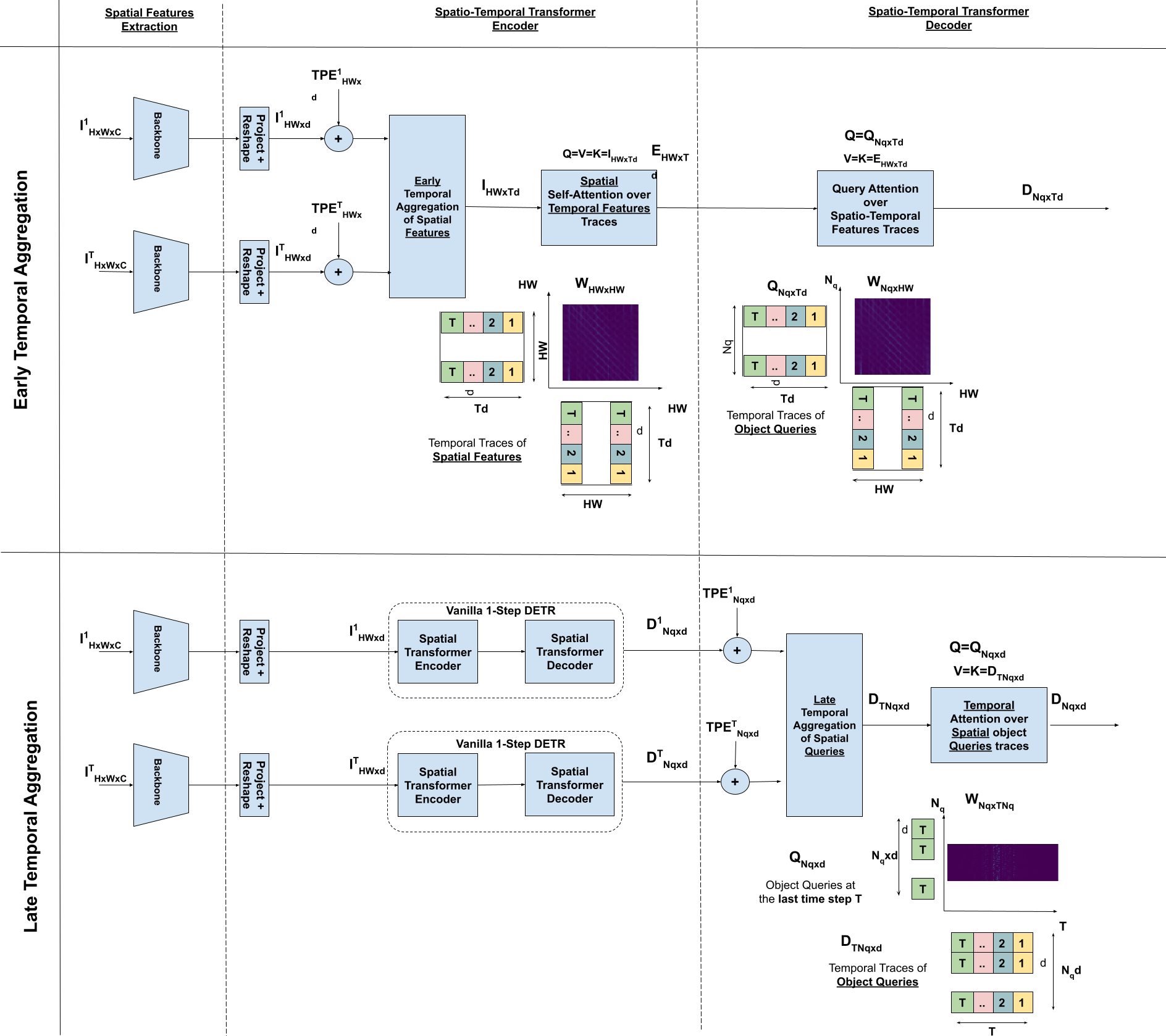}}
\caption{Architectural details of the Early vs. Late Temporal Aggregation variants.}
\label{early_late}
\end{center}
\end{figure*}
\subsection{Spatio-Temporal Detection Transformer}

To transform the vanilla 1-step DETR to deal with temporal sequences, firstly, we have to first deal with multiple streams across $T$ time steps, each having a spatial feature $I_{HW \times d}$, resulting in $I_{HW \times Td}$ streams.
Then using Spatio-Temporal Transformer Encoder (ST-TE) which performs self-attention over the spatial $HW$ dimension, resulting in $E \in \mathbb{R}^{HW \times Td}$.
Finally, by exploiting the Spatio-Temporal Query Transformer Decoder (ST-TD), which performs the query-to-spatial multi-head attention transformation, resulting in $D \in \mathbb{R}^{N_{q} \times d_{final}}$, where $d_{final}$ is the final dimension after spatio-temporal queries aggregation.
The rest of the components remains the same as in vanilla transfomer.

 One can think of two alternatives of temporal features aggregation in both the ST-TE and ST-TD, where we can early aggregate the spatial features over the temporal dimension in the ST-TE, or defer the temporal aggregation to the ST-TD to be done late over the object queries.




\subsubsection{Early Temporal Aggregation}

In this alternative, the list of $T$ spatial features $I_{HWxd}$ are aggregated and flattened into $I_{HWxTD}$. This aggregated tensor $I_{HWxTD}$ can thought of as a spatial map of $T$ temporal traces of spatial features, each of dimension $d$, mapped to the spatial locations $H \times W$. This is visualized in Figure \ref{early_late}. The ST-TE will then perform multi-head self-attention over the spatio-temporal map of object features traces. In this case, we have $Q=V=K=I_{HW \times Td}$. The spatio-temporal features traces attention map $W_{HW \times HW}$\text{=Softmax}$(QK^T)$ is then used to obtain the spatio-temporal features $E_{HW \times Td} = W_{HW \times HW}I_{HW \times Td}$.

The ST-TD will perform multi-head query-to-spatio-temporal features traces attention, where $Q \in \mathbb{R}^{N_{q} \times Td}$ and $V=K=E_{HW \times Td}$. The query-spatio-temporal features traces attention map will be $W_{N_{q} \times HW}$ \text{=Softmax}$(QK^T)$, resulting in $D_{N_q \times Td} = W_{N_{q} \times HW}E_{HW \times Td}$. This represents the final object queries spatio-temporal features, where $d_{final} = Td$ in this case.

\begin{figure*}
\begin{center}
\centerline{\includegraphics[width=170mm]{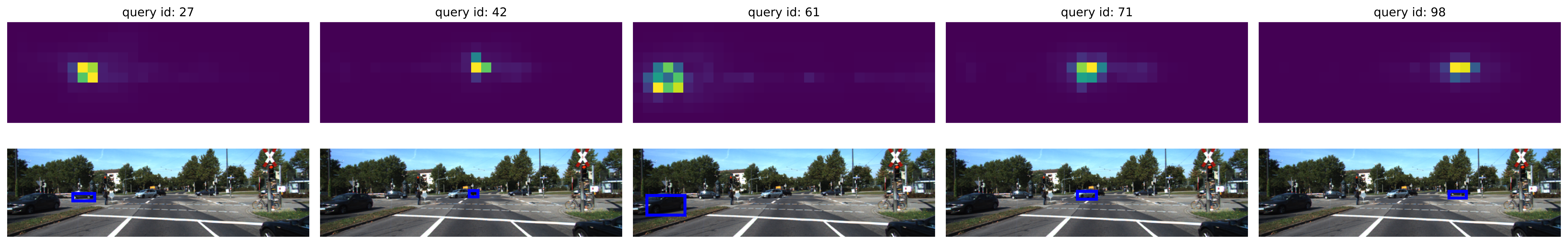}}
\caption{Attention maps for each quires and the corresponding output bounding box .}
\label{attention_map}
\end{center}
\end{figure*}
\subsubsection{Late Temporal Aggregation}
\begin{table}
\caption{Detailed comparisons on the effect of the motion features.}
\label{tab:MODETR}
\begin{center}
\begin{tabular}{lccc}
\hline
Method & $mAP_{Total}$ & $AP_{50}$ & $AP_{75}$ \\
\hline
RGB-only              & 23\% & 42.2\% & 23.7\%                          \\
RGB+RGB           & 25.3\% & 47.2\% & 24.5\%                        \\
RGB + OF    & \textbf{33.9\%} & \textbf{59.3\%} & \textbf{37.2\%}         \\
\hline
\end{tabular}
\end{center}
\end{table}

We could also defer the temporal aggregation until the object queries are obtained per each time step. In this case, the resulting list of $T$ spatial features each of $I_{HW \times d}$ dimension are not stacked and flattened as in the early aggregation.
The ST-TE is formed on $T$ Spatial Transformer Encoders, same as in the vanilla DETR, each performing multi-head self-attention, resulting also in a list of $T$ spatial features $E_{HWxd}$.
Finally, the ST-TD is formed of two levels of decoders; spatial and temporal query decoders.

\textbf{Spatial Query Decoders} which are a list of $T$ decoders, each performing multi-head attention, resulting in a list of $T$ query features each is $D_{N_q \times d}$, which are then reshaped into an aggregated tensor over the temporal dimension to be $D_{T \times N_qd}$. Those represent the Spatio-temporal queries traces.

\textbf{Temporal Query Decoder} which transforms the Spatio-temporal queries traces into the final query features, using multi-head attention. The Spatio-temporal queries traces are first flattened such that $V=K=D_{TN_q \times d}$. The attention learn-able object queries will be $Q \in \mathbb{N_q \times d}$, resulting in an attention map of dimensions $W_{N_q \times TN_q}$\text{=Softmax}$(QK^T)$. This is illustrated in Figure \ref{early_late}. The $TNq$ dimension represents the flattened late $T$ object queries features, each of dimension $d$. This can be thought of as the temporal traces of objects queries as opposed to the objects features traces in the early aggregation alternative. while the $N_q$ dimension represents the final object queries of the last time step, which are to be learned from attending to all the $T$ times steps objects queries. Thus, the final object query features are then $D_{N_q \times d} = W_{N_q \times TN_q}D_{TN_q \times d}$.

\subsection{Sequence-to-sequence prediction}

One can notice from Figure \ref{early_late} that the temporal attention takes place between the last time ($t=T$) step queries; $Q_{N_qxd}$ and the temporal traces of object queries over all the previous steps; $D_{TN_qxd}$. The reason is that we predict the objects in the last frame, given all the features of the previous frames. However, it is straightforward to modify the architecture to obtain a sequence of temporal predictions of $N_q$ object queries per each time step $T$. Simply, in the Temporal Query Decoder, we need to set the queries to $Q_{TNq \times d}$, and thus we have a temporal attention map $W_{TN_q \times TNq}$. This can be thought of as a sequence-to-sequence prediction problem, similar to the Neural Machine Translation (NMT) setup in \cite{vaswani2017attention}, where we have an input sequence of the spatial feature over time, and we predict another sequence of object bounding boxes and classes that corresponds to those inputs.

\begin{table}
\caption{Comparing the Early architecture, Late architecture, and vanilla one step DETR.}
\label{tab:ST-DETR-Early-Late}
\begin{center}
\begin{tabular}{lccc}
\hline
Method & $mAP_{Total}$ & $AP_{50}$ & $AP_{75}$ \\
\hline
1-Step DETR              & 33.9\% & 59.3\% & 37.2\%                          \\
Early           & \textbf{38.7\%} & \textbf{63.1\%} & \textbf{44.6\%}       \\
Late            & 34\% & 61.1\% & 36.1\%         \\
\hline
\end{tabular}
\end{center}
\end{table}
\subsection{Temporal Positional Encoding (TPE)}

Transformers are originally presented as a replacement to recurrent models, due to their fast parallel encoding nature \cite{vaswani2017attention}. However, this comes at the cost of losing the sequential information of the input. To overcome that, positional encoding embedding was proposed in \cite{vaswani2017attention}. Following on that, the vanilla 1-step DETR \cite{carion2020end} treats the input features as being sequential in the spatial dimension $HW$, which leads to the proposal of Spatial Positional Encoding (SPE). In ST-DETR, a similar encoding is needed to distinguish the temporal sequential information of frames. Hence, we propose a Temporal Positional Encoding (TPE), which is added just before the temporal aggregation takes place, being it early across the spatial features traces $TPE_{HWxd}$ or late across the object queries traces $TPE_{Nqxd}$, see Figure \ref{early_late}.
\section{EXPERIMENTS AND RESULTS}

In this section, we first describe the used datasets. After that, we specify the experimental setup, including all hyper-parameters, and hardware specifications. Finally, We design our experiments to evaluate each of our contributions, in the form of an ablation study to evaluate the impact of each one.

\subsection{Dataset}
We use the extended version \cite{rashed2019fusemodnet} of the publicly available KittiMoSeg dataset \cite{siam2018modnet}, that consists of 12919 frames which are split into 80\% for training, and 20\% for testing. The image resolution is $1242 \times 375$, and the labels determine whether the object is moving or static, includes the object bounding box and the motion mask.

\begin{table}
\caption{Quantitative comparison results showing the effect of TPE}
\label{tab:ST-DETR-TPE}
\begin{center}
\begin{tabular}{lccc}
\hline
Method & $mAP_{Total}$ & $AP_{50}$ & $AP_{75}$ \\
\hline
Early              & 36\% & 62.3\% & 43.4\%                          \\
Early+TPE          & \textbf{38.7\%} & \textbf{63.1\%} & \textbf{44.6\%}    \\
\hline
\end{tabular}
\end{center}
\end{table}
\subsection{Experimental Setup}

We initialize our backbone networks with the weights pre-trained on ImageNet \cite{deng2009imagenet}, then train the whole network for 30 epochs on COCO dataset \cite{lin2014microsoft} while freezing the backbone during the first 10 epochs.
In all our experiments, ResNet-50 \cite{he2016deep} was used.
Our network is trained with Adam optimizer \cite{adam} with a scheduled learning rate that is decreased from $1e^{-3}$ to $1e^{-5}$, the whole network is end-to-end trained with learning rate exponentially decayed.
We train a total of 200 epochs, using a warm-up learning rate of $1e^{-3}$ to $5e^{-3}$ in the first 5 epochs.
$460 \times 140$ resolution images have been used across all the experiments.
Our approach is implemented in Python using PyTorch on a PC with Intel Xeon(R) 4108 1.8GHz CPU, 64G RAM, Nvidia Titan-XP.

\subsection{Motion Features}
Previous works on MOD \cite{siam2017modnet, siam2018modnet} indicates that input features can have a strong impact on the results. In particular, features holding motion cues can be of high impact. Thus, we evaluate the best input features at each time step, where we compare RGB, RGB+RGB, and RGB+OF options. In this setup, we use the vanilla 1-step DETR architecture.
Optical flow is generated using FlowNet 2.0 \cite{ilg2017flownet}.
The results are infavor of the RGB+OF setup as shown in Table \ref{tab:MODETR}.

\subsection{Early vs Late temporal aggregation}

In this setup, we evaluate the two architectural alternatives in Figure \ref{early_late}. For the sake of comparison, we fix the time window $T=2$, the number of queries $N_q=100$ and the transformer hidden dimension $d=256$. Results are shown in Table \ref{tab:ST-DETR-Early-Late}. Both results of early and late architectures improve over the 1-step baseline. However, the early architecture provides a significant improvement of 5\% mAP.



\subsection{Effect of TPE}

In this experiment, we evaluate the addition of TPE. Building on the results of early temporal aggregation in Table \ref{tab:ST-DETR-Early-Late}, we perform this comparison on the early temporal setup as shown in Figure \ref{early_late}. As expected, results in Table \ref{tab:ST-DETR-TPE}, show 2\% mAP improvement over the variant without TPE.

\subsection{Effect of the temporal window size T}
\label{Effect of the temporal window size T}

In this setup, we evaluate the effect of the increased window size, for $T=1,2,4$. Results in Table \ref{tab:ST-DETR-T} show increased performance with the increase of $T$. However, a saturation barrier is hit at $T=4$.

\begin{table}
\caption{Quantitative comparison results showing the effect of the temporal window size T}
\label{tab:ST-DETR-T}
\begin{center}
\begin{tabular}{lccc}
\hline
T & $mAP_{Total}$ & $AP_{50}$ & $AP_{75}$ \\
\hline
1-Step              & 33.9\% & 59.3\% & 37.2\%         \\
2-Steps          & \textbf{38.7\%} & 63.1\% & \textbf{44.6\%}      \\
4-Steps          & \textbf{38.7\%} & \textbf{64.8\%} & 43\%   \\
\hline
\end{tabular}
\end{center}
\end{table}

\section{CONCLUSION}
In this work, we extend the vanilla DETR architecture, into a Spatio-Temporal model to deal with video inputs. We explore various design choices in our endeavor; the early vs. late temporal aggregation setups. Results are in favor of the early architecture which deals with temporal traces of spatial motion features. Our analysis of the 1-step motion features suggests that the best option is to feed the RGB+OF frames of the input 1-step scene, which is also in line with previous works. We also propose an extra Temporal Positional Embedding (TPE) step, to enable the temporal differentiation of features. Results show improved performance with TPE introduced to the architecture. The new ST-DETR architecture achieves 5\% mAP improvement on the KITTI MOD dataset.

\bibliography{main.bib}
\bibliographystyle{icml2021}
\end{document}